\pdfoutput=1
\documentclass{article}

\usepackage[table,dvipsnames]{xcolor}
\usepackage[preprint]{corl_2026} 

\usepackage{xspace}
\usepackage{algorithmicx}
\usepackage{algpseudocode}
\usepackage{amsmath}
\usepackage{multirow}
\usepackage{makecell}
\usepackage{bm}
\usepackage{booktabs}
\usepackage{graphicx}
\usepackage{bbding}
\usepackage{amssymb}
\usepackage{tikz}
\usepackage{hyperref}
\usepackage[capitalise]{cleveref}
\crefname{figure}{Fig.}{Figs.}
\Crefname{figure}{Fig.}{Figs.}

\crefname{table}{Tab.}{Tabs.}
\Crefname{table}{Tab.}{Tabs.}
\usepackage{caption}
\usepackage{enumitem}
\usepackage[normalem]{ulem}
\usepackage{subcaption}
\usepackage[most]{tcolorbox}
\usepackage{titletoc}

\newcommand{\ourmethod}{UniLM-Nav\xspace}

\definecolor{myblue_view_selection}{HTML}{4C72B0}
\definecolor{mygreen_base_pose}{HTML}{55A868}
\definecolor{myred_nav_adjustment}{HTML}{C44E52}
\definecolor{myred_error_analysis}{HTML}{FF0000}
\definecolor{mygreen_error_analysis}{HTML}{00B050}
\definecolor{myblue_project_url}{HTML}{3B6EA8}

\title{\ourmethod: A Unified Framework for Zero-Shot Last-Mile Navigation}

\author{%
  Zhuofan Zhang$^{1,2*}$, Tianxu Wang$^{2*}$, Guoxi Zhang$^{2}$, Yixiong Lin$^{2,3}$, \\
  \bf Xilin Wang$^{2}$, Hongming Xu$^{2}$, Qing Li$^{2}$, Song-Chun Zhu$^{1,2,4}$, Lifeng Fan$^{2\dagger}$ 
  \vspace{0.5em}\\
  $^{1}$Tsinghua University \\
  $^{2}$State Key Laboratory of General Artificial Intelligence, BIGAI \\ 
  $^{3}$Harbin Institute of Technology \\
  $^{4}$Peking University\\
}

\makeatletter
\def\thanks#1{\protected@xdef\@thanks{\@thanks \protect\footnotetext{#1}}}
\makeatother

\thanks{$^{*}$Equal contribution. $^{\dagger}$Corresponding author.}

\begin{document}
\maketitle

\vspace{-2.1em}
\centerline{
  \textbf{
    Project page:
    \href{https://unilm-nav.github.io}
    {\color{myblue_project_url}https://unilm-nav.github.io}
  }
}
\vspace{1.2em}


\begin{abstract}
Mobile manipulation requires a robot to navigate to a target object or receptacle and then perform intended manipulation.
However, reaching the vicinity of the target does not guarantee a manipulation-ready base pose, a problem known as \textbf{last-mile navigation}.
Prior methods for last-mile navigation either rely on manual pose annotation or task-specific training, limiting their scalability to open-vocabulary settings with fine-grained spatial constraints.
We propose \textbf{\ourmethod}, a \underline{uni}fied framework for zero-shot open-vocabulary \underline{l}ast-\underline{m}ile \underline{nav}igation. \ourmethod decomposes last-mile navigation into view selection, task-conditioned affordance grounding, and geometry-aware base-pose reasoning, all resolved with a shared multimodal large language model (MLLM) backend.
Specifically, \ourmethod first selects a reference view that best captures the target object or receptacle from recently collected observations. 
It then grounds task-relevant affordance point in the selected view and lifts the result into the robot-centric coordinate frame.
Finally, conditioned on the grounded affordance, task context, and robot geometry, it infers a manipulation-ready base pose for the robot.
We evaluate \ourmethod on the OVMM benchmark, where it outperforms the previous state-of-the-art method, MoTo, by 3.13 percentage points. Analyses show that the components of our method are crucial to final performance, and that the choice of MLLM also has a substantial effect.
We further deploy \ourmethod on a Unitree B2 quadruped robot with a 6-DoF Unitree Z1 manipulator, validating its applicability to real-world mobile manipulation tasks.

\end{abstract}

\keywords{Mobile Manipulation, Last-Mile Navigation, MLLM} 


\section{Introduction}
\label{sec:intro}

    \begin{figure}[t]
      \centering
      \includegraphics[width=0.97\linewidth]{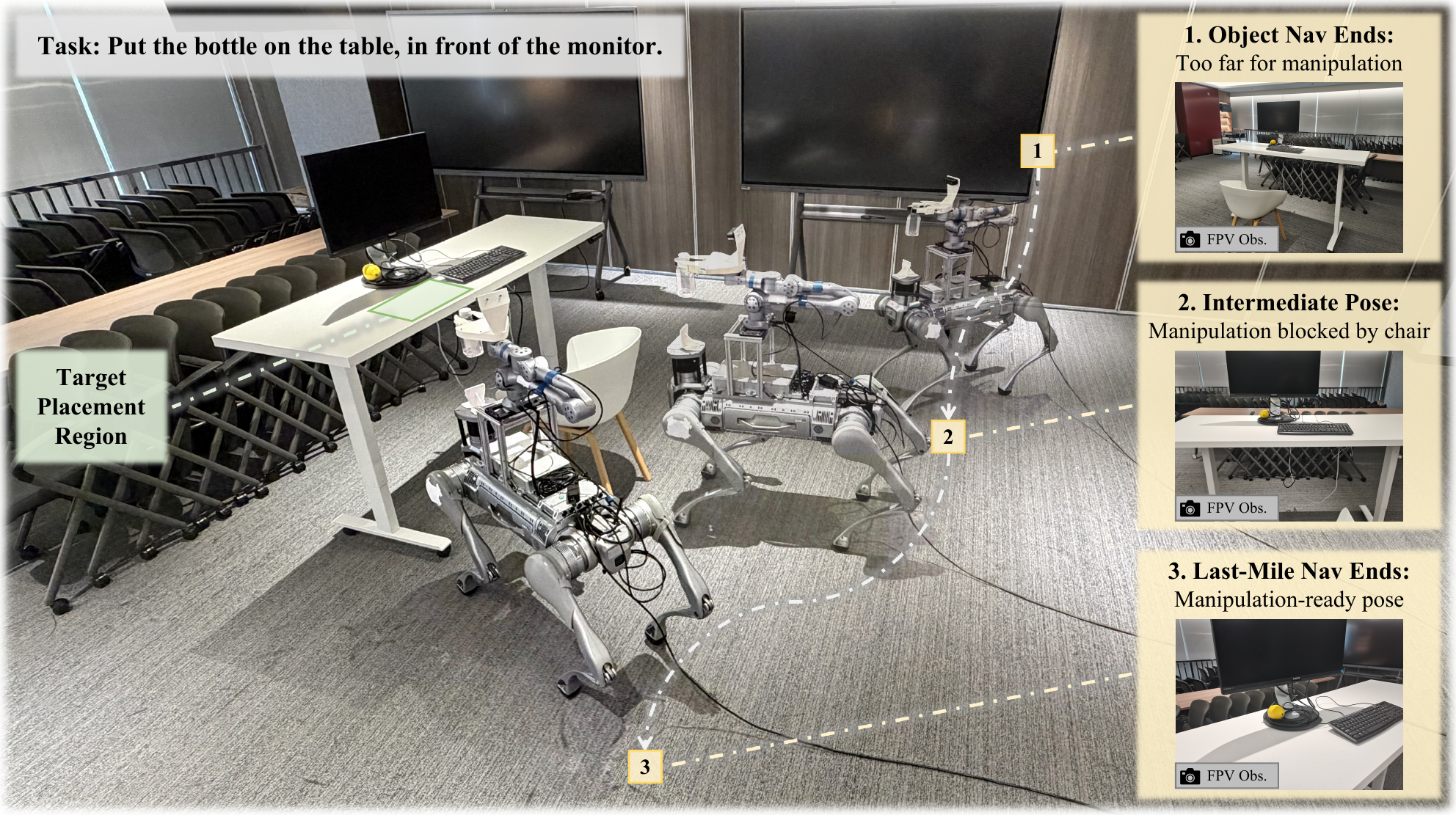}
      \caption{
    The robot is tasked with placing the bottle on the table in front of the monitor. Object-goal navigation can bring the robot near the table, but the resulting base pose may still be infeasible for placement due to limited reachability or surrounding obstacles. Last-mile navigation refines its base pose based on the task context and robot-scene geometry, thereby enabling feasible placement.
        }
      \label{fig:teaser}
      \vspace{-1.5em}
    \end{figure}

Mobile manipulation requires robots to navigate to task-relevant objects or regions and manipulate objects~\cite{harmonic-mobile-manipulation, momanipvla}, a core capability for household service~\cite{robi-butler, composable-interaction-primitives}, industrial automation~\cite{revolutionizing-battery-disassembly}, and logistics~\cite{mmforlogistics}.
In mobile manipulation, the navigation phase should end in a \textit{manipulation-ready base pose}---a pose that is not only close to the target but also supports the intended manipulation.
However, object navigation systems are typically designed to reach the vicinity of the target~\citep{goat}, e.g., within 1--2 meters, which is coarser than manipulation requires.
This granularity mismatch can leave the robot poorly aligned or obstructed for manipulation.
This motivates the task of \textbf{last-mile navigation}~\citep{moma-kitchen}: once near the target, the robot must adjust its base pose to support the manipulation.


Last-mile navigation requires jointly identifying interaction-relevant affordance regions and selecting a feasible base pose under spatial and kinematic constraints.
Early approaches rely on manual annotation for task-specific base poses~\citep{wildlma}, limiting their generalization to new tasks or environments.
Alternatively, learning-based methods either handle last-mile navigation implicitly with monolithic navigation-manipulation policies~\citep{skill-transformer,asc} or explicitly learn manipulation-conditioned pose preferences~\citep{moma-kitchen, mobi-pi}, but they remain data-intensive and limited in open-vocabulary settings.
Thus, there is still a need for a flexible, training-free method in open-vocabulary settings.

Recently, vision foundation models have advanced open-vocabulary perception~\citep{sam, dinov2, clip}, while multimodal large language models (MLLMs) have demonstrated visual reasoning abilities~\citep{gpt4o,o3o4mini}.
These capabilities have inspired pipelines that use object-level cues to reason about base poses for last-mile navigation~\citep{moto,affordance_rgb}.
However, object-level reasoning is insufficient for satisfying fine-grained spatial constraints.
Consider the example task shown in~\cref{fig:teaser}, where the task is to put a bottle in front of the monitor. 
The task is not merely to localize the monitor, but to identify a receptacle region satisfying a particular spatial relation.
Therefore, it remains an open question how last-mile navigation can move beyond object-level cues toward task-aware spatial reasoning.

Leveraging the spatial reasoning capabilities of MLLMs~\citep{GPT5_SI, qwen3-vl,gemini-3-flash}, we propose \textbf{\ourmethod}, a zero-shot MLLM framework for inferring manipulation-ready base poses through task-aware spatial reasoning.
\ourmethod decomposes last-mile navigation into three phases: view selection, task-conditioned affordance grounding, and geometry-aware base-pose reasoning.
Firstly, \ourmethod builds a lightweight observation memory and uses the MLLM to select a reference view suitable for affordance grounding.
It then grounds the task-relevant region in the selected view.
Finally, \ourmethod lifts the grounded 2D location into the robot-centric 3D frame and uses the same MLLM to infer an appropriate base pose.
Compared with prior MLLM-based approaches~\citep{moto,affordance_rgb}, \ourmethod offers a more unified and implementation-friendly formulation by using a shared MLLM backend.

We evaluate \ourmethod using the OVMM benchmark~\citep{homerobot}.
\ourmethod, with Gemini-3-Flash-Preview as backend, surpasses MoTo~\citep{moto}, the previous state-of-the-art (SOTA) method, by 3.13 percentage points in overall success rate.
We further conduct empirical analyses of the components of \ourmethod, the choice of MLLM backend, and its failure modes.
Interestingly, with a 4B MLLM backend, RoboBrain-2.5-4B~\citep{robobrain2_5} outperforms variants using proprietary models such as GPT-5.4~\citep{gpt5.4} and larger open-source models such as Qwen3-VL-235B-A22B-Instruct~\citep{qwen3-vl}, probably due to its fine-tuning on robotic data.
Finally, we deploy \ourmethod on a Unitree B2 quadruped robot~\citep{unitree_b2} with a 6-DoF Unitree Z1 manipulator~\citep{unitree_z1} and validate it on four real-world tasks.
To summarize, our contributions are as follows.
    \begin{itemize}[leftmargin=10pt]
        \item We propose \ourmethod, a unified zero-shot MLLM framework that employs task-aware spatial reasoning for last-mile navigation.
        \item We show that \ourmethod achieves SOTA performance on the OVMM benchmark and validate it on real robots.
        \item We provide empirical analyses of framework components, MLLM backends, and failure modes, offering insights for designing MLLM-based last-mile navigation systems.
    \end{itemize}


\section{Related Works}
\label{sec:Related Works}
\noindent\textbf{Mobile Manipulation}\quad Existing approaches for mobile manipulation can be broadly divided into end-to-end methods and modular methods.
End-to-end methods~\citep{rt-1,mobile_aloha,m2diffuser} learn a joint policy for base and arm actions from large-scale demonstrations, which is costly to scale to open-vocabulary and long-horizon tasks.
Modular approaches~\citep{homerobot,ok-robot,nav_abstract} decouple navigation and manipulation, improving data efficiency and scalability.
These methods often adopt heuristics to determine manipulation-ready base poses, including pose scoring~\citep{ok-robot}, sampling-based region estimation~\citep{nav_abstract}, feasibility predicates~\citep{ialp}, trajectory back-projection~\citep{momanipvla}, or motion planning~\citep{limp}.
While these strategies help bridge navigation and manipulation, their reliance on predefined heuristics can limit flexibility under open-vocabulary tasks with fine-grained spatial constraints.

\noindent\textbf{Last-Mile Navigation}\quad Prior work on last-mile navigation differs primarily in how manipulation-ready base poses are obtained.
A class of methods learns a policy for last-mile navigation using imitation learning~\citep{last_meter_navigation} or reinforcement learning~\citep{aprr}.
MoMa-Kitchen~\citep{moma-kitchen}, instead, directly learns distributions over manipulation-ready base poses from large-scale supervision.
Another thread of work models base pose preferences conditioned on downstream manipulation policies~\citep{mobi-pi,n2m}.
Recent approaches like MoTo~\citep{moto} and~\citep{affordance_rgb} leverage vision foundation models and MLLMs for zero-shot open-vocabulary generalization. While effective for interaction-centric navigation, these methods leave task-aware spatial-relation reasoning under-explored.



\noindent\textbf{MLLMs for Embodied Control}\quad Prior applications of MLLMs in embodied control have mainly focused on affordance grounding and manipulation.
PIVOT~\citep{pivot} and MOKA~\citep{moka} leverage MLLMs to localize task-relevant keypoints, candidate regions, or trajectories, whereas AffordGrasp~\citep{affordgrasp} uses MLLMs to reason about grasp poses.
Meanwhile, recent studies have evaluated and improved MLLM spatial reasoning through spatial benchmarks, spatially grounded data, and geometry-aware supervision~\citep{VSI,spatialvlm,omnispatial,theory_of_space,spatialmllm}.
Such capabilities have been further explored in embodied tasks including task planning~\citep{embodiedbench}, affordance referring~\citep{roborefer}, and embodied spatiotemporal reasoning~\citep{robobrain2_5}.
However, MLLM's application to last-mile navigation remains largely under-explored, despite the need for visual-spatial reasoning over affordances, free space, obstacles, and spatial relations.



\section{Problem Statement}
In open-vocabulary mobile manipulation (OVMM), a robot is given a natural-language instruction specifying a task that requires navigation with object manipulation.
We assume that the robot is equipped with an object-navigation policy and a pre-built scene map, which together allow it to move to the vicinity of the target object or receptacle. 
This object-navigation stage is considered complete once the target has been observed and lies within a predefined near-target radius, e.g., 1--2 meters.
However, proximity to the target alone does not guarantee successful manipulation, since the robot's current base pose may not provide the reachability, orientation, or clearance required for the intended manipulation.
We refer to the problem of positioning the robot in a manipulation-ready base pose after object navigation as \emph{last-mile navigation}.

During last-mile navigation, the robot has access to egocentric RGB-D observations, its proprioceptive state, odometry estimates, task instructions, and an obstacle map estimated from onboard sensors. 
Given these inputs, the goal is to select a manipulation-ready base pose $\mathbf{b}=(x,y,\theta) \in \mathbb{R}^3$, where $(x,y)$ denotes the target base position and $\theta$ denotes the desired heading. 
The selected pose should be collision-free, reachable, and suitable for the intended manipulation. 
In this work, we study how an off-the-shelf MLLM can be used to reason about this pose in a zero-shot manner.


\section{Method}
\label{sec:Method}
\begin{figure}
    \centering
    \includegraphics[width=\linewidth]{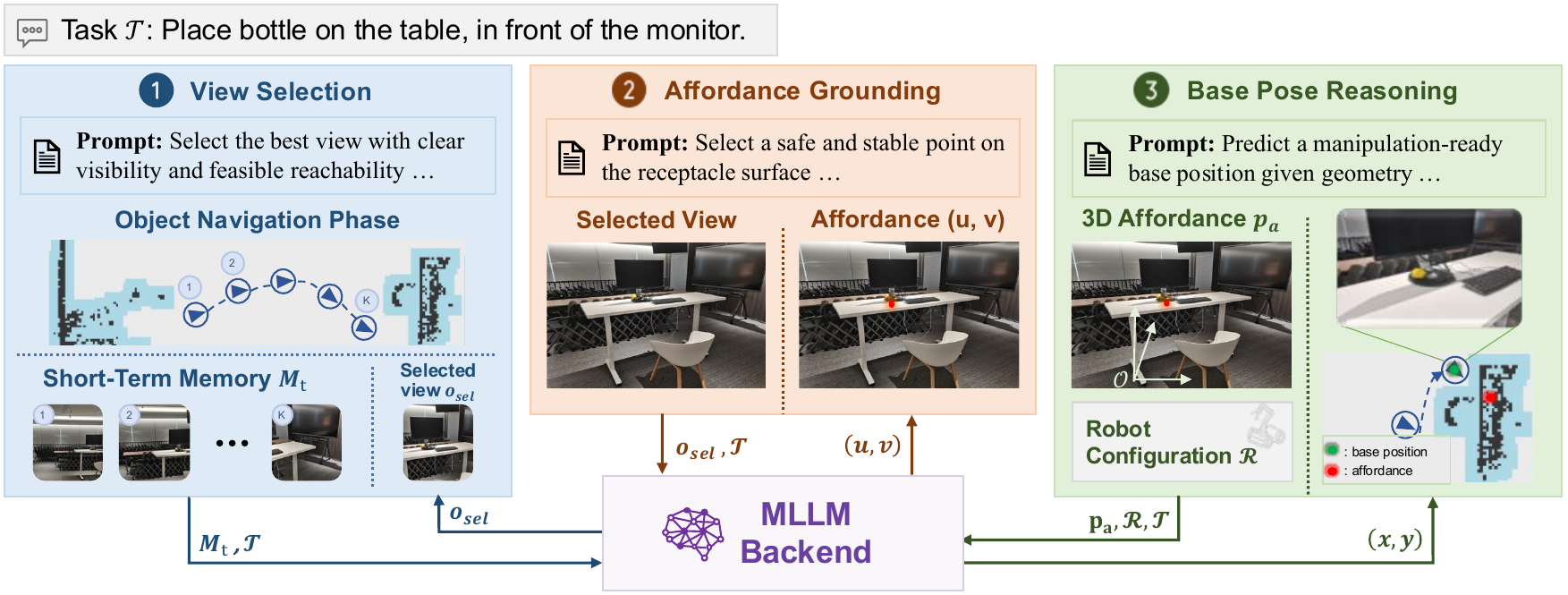}
    \caption{\ourmethod decomposes last-mile navigation into view selection, affordance grounding, and base-pose reasoning. Given a task instruction, the MLLM first selects a suitable view from recent observations, then grounds the task-relevant affordance and lifts it to 3D with depth. Conditioned on the 3D affordance, robot configuration, and task instruction, the MLLM predicts a manipulation-ready base pose for execution.}
    \label{fig:method_teaser}
    \vspace{-1em}
\end{figure}


We propose \ourmethod, a zero-shot MLLM framework for last-mile navigation that bridges object navigation and manipulation-ready positioning. As illustrated in~\cref{fig:method_teaser}, \ourmethod decomposes this process into three stages: view selection, task-conditioned affordance grounding, and geometry-aware base-pose reasoning.


\noindent\textbf{View Selection.}\quad
Upon completion of object-goal navigation, the robot is expected to be in the vicinity of the target object or receptacle, typically within about 1--2 meters.
However, the final observation captures only a single view along the recent navigation trajectory and may not provide the most informative view for manipulation.
The target object or receptacle surface may be partially occluded, or the surrounding layout may not provide sufficient spatial context for manipulation.
To obtain a more reliable visual basis for last-mile navigation, we maintain a short-term memory buffer over the last $K$ steps before object-goal navigation terminates, defined as:
\begin{equation}
   \mathcal{M}_t=\{(o_{t-k}, s_{t-k})\}_{k=0}^{K-1}, 
\end{equation}
where $t$ denotes the final timestep of the object-goal navigation phase, $K$ is the buffer size, and each entry consists of an egocentric observation $o_{t-k}$ and the corresponding robot state $s_{t-k}$, which specifies the robot and camera poses.
Given these candidate observations and task instruction, we prompt the MLLM to select a reference view $o_{\mathrm{sel}}$ that best supports downstream manipulation, according to two criteria:
whether the target object or receptacle is clearly visible, and whether the robot has a feasible approach path to navigate close enough for manipulation. The selected observation then serves as the reference view for subsequent task-conditioned affordance grounding and geometry-aware base-pose reasoning.

\noindent\textbf{Task-Conditioned Affordance Grounding.}\quad
Given the selected observation $o_{\mathrm{sel}}$, we further ask the MLLM to ground the task instruction to a concrete image-space affordance point. 
Unlike object-level localization, this step requires the model to identify the specific scene location where the robot should interact to accomplish the task.
Formally, given the selected observation $o_\mathrm{{sel}}$ and task instruction $\mathcal{T}$, the MLLM predicts an image-space affordance point: 
\begin{equation}
    (u, v) = \mathrm{MLLM}(o_{\mathrm{sel}}, \mathcal{T}),
\end{equation}
where $(u,v)$ denotes the predicted 2D pixel location at which the intended manipulation should be performed. 
For pick tasks, this point is expected to lie in a graspable region of the target object.
For placing tasks, it is expected to correspond to a safe and unoccupied placement location on the surface of the target receptacle.
This step requires the MLLM to jointly reason about the task instruction, object semantics, local spatial layout, and manipulation constraints.

\noindent\textbf{Geometry-Aware Base-Pose Reasoning.}\quad
Given the grounded image-space affordance point, a straightforward approach to manipulation-ready base-pose prediction is to directly ask the MLLM to select a floor point in the image as the target base position. However, this requires the model to visually infer metric distances, reachability, and geometric constraints from a 2D observation, which remains challenging for current MLLMs.
Rather than relying on implicit visual geometry estimation, we convert visual evidence into explicit geometric quantities and provide them to the MLLM, allowing it to leverage its reasoning capability for base-position prediction.

Specifically, given the 2D affordance point $(u,v)$ predicted in the selected observation $o_{\mathrm{sel}}$, we use the aligned depth map and camera parameters to lift this point into a 3D affordance point $\mathbf{p}_a$ in the coordinate frame of the selected robot pose. We also visually mark the affordance point on the selected observation, resulting in a visually prompted observation $\tilde{o}_{\mathrm{sel}}$.

Then, we provide the MLLM with the visually prompted observation $\tilde{o}_{\mathrm{sel}}$, the robot configuration $\mathcal{R}$, the task instruction $\mathcal{T}$, and the lifted 3D target coordinate $\mathbf{p}_a$.
Conditioned on these inputs, the MLLM predicts a manipulation-ready base position $(x, y)$ in a local coordinate frame centered at the robot pose corresponding to the selected observation. Formally:
\begin{equation}
    (x,y)=\mathrm{MLLM}(\tilde{o}_{\mathrm{sel}}, \mathbf{p}_a, \mathcal{R}, \mathcal{T}),
\end{equation}
Rather than asking the MLLM to directly predict the heading angle $\theta$, we compute the final orientation geometrically by orienting the robot toward the lifted affordance point, resulting in the full local base pose $(x, y, \theta)$. 
We then transform the local base pose into the global frame using the robot pose of the selected observation and pass it to the low-level navigation policy, which drives the robot to the target pose while avoiding collisions.



\section{Experiments}
\label{sec:Experiments}

\begin{table*}[t]
\caption{\textbf{\ourmethod achieves the best performance on the OVMM validation set.}
With a lightweight RoboBrain-2.5-4B as its MLLM backend, \ourmethod still outperforms most alternative methods.
Average SR is computed by averaging the partial success rates and the Overall SR.
}
\centering
\small
\resizebox{\linewidth}{!}{
\begin{tabular}{@{}lcccccc@{}}
\toprule
\multirow{2}{*}{\textbf{Method}}
& & \multicolumn{3}{c}{\textbf{Partial Success Rates}} &\multirow{2}{*}{\textbf{\makecell{Overall
\\ SR}}} &\multirow{2}{*}{\textbf{\makecell{Average
\\ SR}}} \\
\cmidrule{3-5} 
  & & FindObj~($\uparrow$) & Pick~($\uparrow$) & FindRec~($\uparrow$)  \\
 \midrule
HomeRobot (RL)~\citep{homerobot} &  &66.60\%  & 61.10\% &  50.90\% & 14.80\% & 48.30\%  \\
HomeRobot (Heuristic)~\citep{homerobot} &  &65.40\%  &54.80\% &43.70\% &7.30\% &42.80\%   \\
MoManipVLA~\citep{momanipvla} & &66.10\%  &62.60\%  &53.10\%  &15.80\%  &49.40\%  \\
UniTeam~\citep{uniteam} &  &66.13\% &62.65\%  &54.69\% &  17.96\% & 50.36\%   \\
MoTo~\citep{moto} & &66.67\% &60.95\% &49.87\% &20.64\% &49.53\%\\
\midrule
 \rowcolor{gray!20}
 \ourmethod (w/ Gemini-3-Flash-Preview) & &69.47\% &66.22\% &54.55\% &23.77\% &53.50\% \\
  \rowcolor{gray!20}
 \ourmethod (w/ RoboBrain-2.5-4B) & &68.97\% &63.05\% &52.38\% &19.19\% &50.90\% \\

 \bottomrule
\end{tabular}
}
\label{table: main results}
\vspace{-1em}
\end{table*}

\subsection{Evaluations on OVMM Benchmark}
\textbf{Setup}\quad We evaluate \ourmethod on the HomeRobot OVMM benchmark~\citep{homerobot}, where a robot is required to find a target object, pick it up, navigate to a target receptacle, and place the object in an open-vocabulary indoor environment.
Following the official evaluation protocol, we report the overall success rate (Overall SR) together with cumulative stage success rates up to FindObj, Pick, and FindRec, where each stage is evaluated conditioned on the successful completion of the preceding stages.
We adopt Gemini-3-Flash-Preview~\citep{gemini-3-flash} as the MLLM backend of \ourmethod, due to its favorable balance between response latency and multimodal reasoning capability.
For navigation and grasping, we adopt the default HomeRobot policies. For placement, we implement a simple manipulation policy integrated with our base-pose reasoning stage. In addition to the target base position, the MLLM predicts the arm extension and lift height required for placement, which are executed by the robot controller during placement.

\textbf{Baselines}\quad We compare \ourmethod against representative baselines, including the RL-based and heuristic variants of HomeRobot~\citep{homerobot}, the training-based method MoManipVLA~\citep{momanipvla}, and the training-free MoTo approach~\citep{moto}.



\textbf{Main Results} \quad As shown in~\cref{table: main results}, \ourmethod (w/ Gemini-3-Flash-Previw) achieves state-of-the-art performance on the OVMM validation set~\citep{homerobot}, reaching 23.77\% Overall SR.
Notably, despite being fully zero-shot, it outperforms both the training-based method MoManipVLA (15.80\%) and the strongest zero-shot baseline MoTo (20.64\%). Moreover, using the lightweight RoboBrain-2.5-4B as the MLLM backend also achieves a competitive Overall SR of 19.19\%, surpassing most compared baselines and offering a practical trade-off between deployability, inference efficiency, and task performance. In addition to overall success rate, \ourmethod also achieves the highest FindObj success rate of 69.47\% and 68.97\%, indicating that the proposed last-mile navigation strategy enables more reliable target approach after object navigation. 
\subsection{Ablation Study}

\begin{figure*}[t]
  \centering

  \begin{subfigure}[t]{0.43\linewidth}
    \centering
    \includegraphics[width=\linewidth]{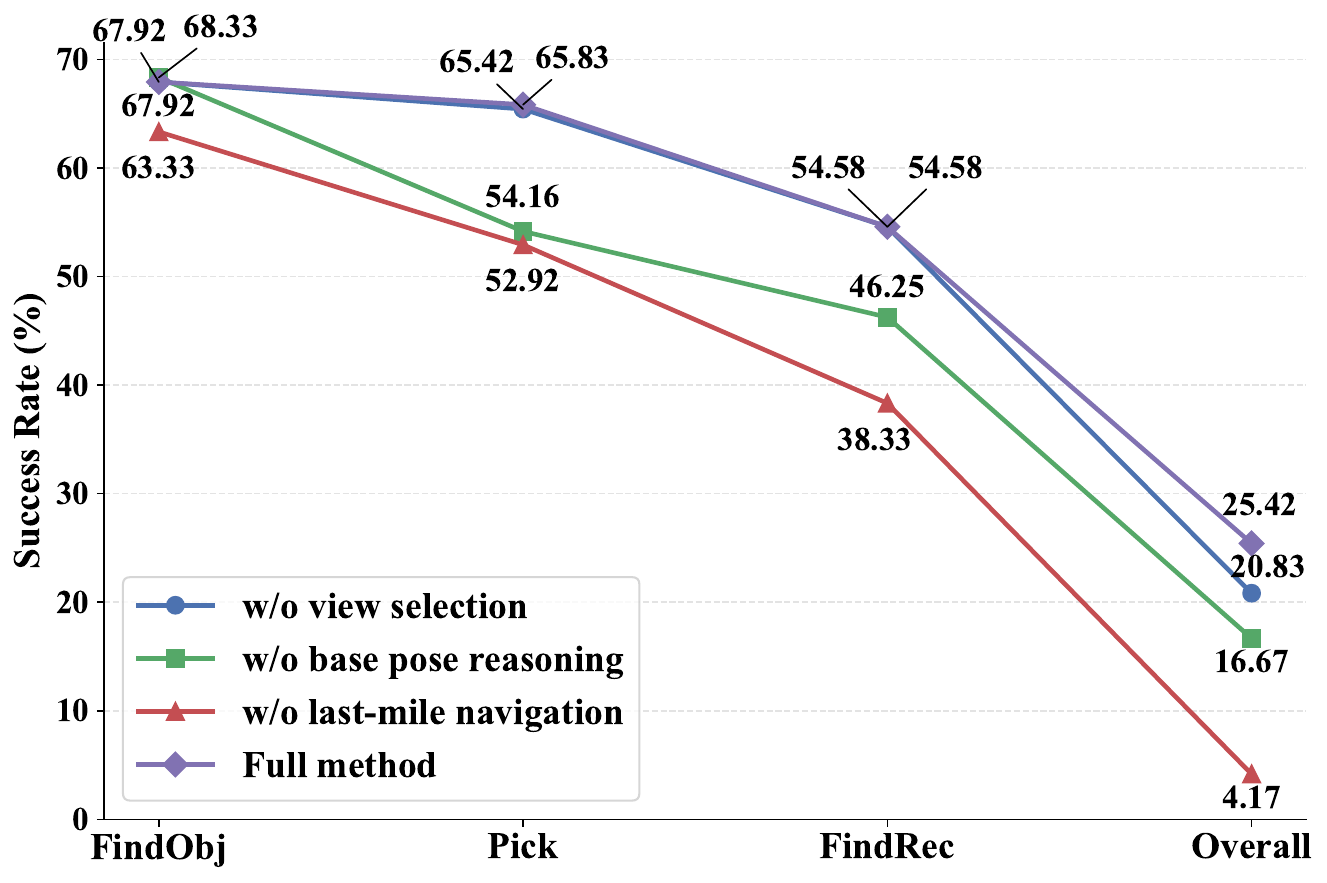}
    \caption{}
    \label{fig:framework_ablation}
  \end{subfigure}
  \hfill
  \begin{subfigure}[t]{0.56\linewidth}
    \centering
    \includegraphics[width=\linewidth]{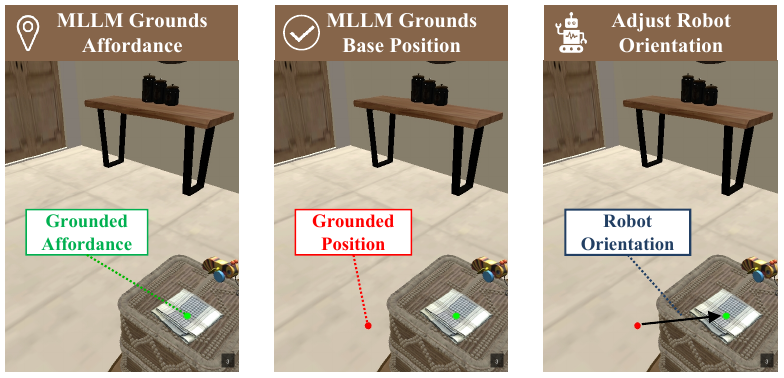}
    \caption{}
    \label{fig:wo_base_pose_reasoning}
  \end{subfigure}

  \caption{
  Framework ablation of \ourmethod.
   (a) Success rates of \ourmethod and its variants on the 20\% subset of the OVMM benchmark. 
  (b) Visualization of the \textcolor{mygreen_base_pose}{w/o base-pose reasoning} variant, where the MLLM directly grounds a floor point as the manipulation-ready base position instead of reasoning over explicit geometry-aware context for base-pose prediction.
  }
  \label{fig:ablation_analysis}
  \vspace{-1.5em}
\end{figure*}

\noindent\textbf{Setup.}\quad
All ablations are conducted on a scene-stratified random 20\% subset of the OVMM benchmark due to the high computational cost of embodied evaluation.
For \textbf{\textcolor{myred_nav_adjustment}{w/o last-mile navigation}}, we skip the last-mile navigation that moves the robot to the predicted manipulation-ready base position. Instead, the robot only turns to face the grounded affordance point before manipulation.
For \textbf{\textcolor{myblue_view_selection}{w/o view selection}}, we remove the MLLM-based view selection module and directly use the final observation from the FindObj and FindRec phases for affordance grounding. 
For \textbf{\textcolor{mygreen_base_pose}{w/o base-pose reasoning}}, we remove the geometry-aware step that predicts a manipulation-ready base position conditioned on the lifted robot-centric affordance point. Instead, the MLLM is prompted to visually infer the target base position by directly grounding a floor point in the image. For \textbf{MLLM backend ablation}, we report results for alternative proprietary, open-source, and embodied MLLMs.

\noindent\textbf{\textcolor{myred_nav_adjustment}{Necessity of Last-Mile Navigation.}}\quad
Removing last-mile navigation reduces the overall SR to drop below 5\%, showing that simply turning the robot toward the grounded affordance point is far from sufficient for successful manipulation.
Even when the target object or receptacle is visible after object navigation, the robot often still needs to move to a physically executable base pose, confirming the necessity of last-mile navigation in OVMM tasks.

\noindent\textbf{\textcolor{myblue_view_selection}{Effectiveness of View Selection.}}\quad
Removing the view-selection module decreases the overall SR from 25.42\% to 20.42\%. This indicates that the final observation after object navigation is not always the most suitable view for downstream manipulation. 
By selecting a more suitable observation from memory, \ourmethod provides clearer visual evidence for affordance grounding and base-pose reasoning, improving the reliability of subsequent manipulation execution.

\noindent\textbf{\textcolor{mygreen_base_pose}{Effectiveness of Base-Pose Reasoning.}}\quad
The variant without base-pose reasoning substantially underperforms \ourmethod in both Pick success and Overall SR. 
This indicates that directly grounding a manipulation-ready floor point in the image is unreliable, as the MLLM must implicitly infer metric distance, reachability, and local feasibility from 2D visual evidence. In contrast, \ourmethod provides explicit robot-centric geometric context for base-pose prediction, enabling more reliable manipulation-ready positioning.

\noindent\textbf{{Effect of MLLM backend choice.}}\quad
As shown in~\cref{table:method ablation}, the choice of MLLM backend leads to noticeable performance variations on the OVMM benchmark. Proprietary models, such as Gemini-3-Flash-Preview and GPT-5.4, achieve relatively strong results. Within the Qwen3-VL-Instruct family, larger models generally achieve higher Overall SR, increasing from 3.75\% for Qwen3-VL-4B to 15.83\% for Qwen3-VL-32B. More importantly, RoboBrain-2.5-4B, which is built on the Qwen3-VL architecture and further adapted with robotics-oriented embodied spatial reasoning data, achieves 20.50\% Overall SR, substantially outperforming Qwen3-VL-4B and even surpassing the much larger Qwen3-VL-235B-A22B-Instruct.
These results suggest that both model scaling and embodied spatial training contribute to improved OVMM performance. 


\begin{table*}[t]

\caption{\textbf{The performance difference of MLLM backends centers at the placement stage.}
On the 20\% subset of the OVMM validation set, all MLLM backends have similar performance for the FindObj stage, while Proprietary models become advantageous in later stages.
Significant performance variations are observed from the FindRec stage to overall task completion, indicating that the placement stage is very challenging for the models. 
Probably due to fine-tuning on robotic data, RoboBrain-2.5-4B attains very competitive performance.}
\centering
\small
\resizebox{\linewidth}{!}{
\begin{tabular}{@{}lcccccc@{}}
\toprule
\multirow{2}{*}{\textbf{Method}}
& & \multicolumn{3}{c}{\textbf{Partial Success Rates}} &\multirow{2}{*}{\textbf{\makecell{Overall
\\ SR}}} &\multirow{2}{*}{\textbf{\makecell{Average
\\ SR}}} \\
\cmidrule{3-5} 
  & & FindObj~($\uparrow$) & Pick~($\uparrow$) & FindRec~($\uparrow$)  \\
 \midrule
\rowcolor{gray!20} \textit{Proprietary MLLM} & & & & & & \\
GPT-4.1~\citep{gpt4.1} &  &68.75\%  &62.50\% &49.58\% &15.42\% &49.06\%   \\
GPT-5.2~\citep{gpt5.2} &  &68.33\%  &65.42\% &54.58\% &18.75\% &51.77\%   \\
GPT-5.4~\citep{gpt5.4} &  &68.33\%  & 66.25\% &  55.00\% & 19.17\% & 52.19\%  \\
Gemini3-Flash-Preview~\citep{gemini-3-flash} & &67.91\% &65.83\% &54.58\% &25.42\% &53.43\% \\
\midrule
\rowcolor{gray!20} \textit{Open-Source MLLM} & & & & & & \\
Qwen3-VL-4B-Instruct~\citep{qwen3-vl} & &68.33\%  &61.25\%  &50.42\%  &3.75\%  &45.93\%  \\
Qwen3-VL-8B-Instruct~\citep{qwen3-vl} & &68.75\%  &60.00\%  &47.92\%  &9.58\%  &46.56\%  \\
Qwen3-VL-32B-Instruct~\citep{qwen3-vl} &  &67.92\% &62.08\%  &52.08\% &  15.83\% & 49.48\%   \\
Qwen3-VL-235B-A22B-Instruct~\citep{qwen3-vl} &  &67.92\% &59.58\%  &48.33\% &  17.50\% & 48.33\%   \\
Qwen3.5-27B~\citep{qwen3.5} &  & 68.33\% & 64.17\%  &  54.58\% &  18.33\% & 51.35\%   \\
Qwen3.6-27B~\citep{qwen3.6} &  &67.50\% & 65.00\%  &54.58\% &  20.00\% & 51.77\%   \\
InternVL3.5-8B~\citep{internvl3_5} &  &66.67\% & 50.00\%  &40.00\% &  8.75\% & 41.35\%   \\
InternVL3.5-14B~\citep{internvl3_5} &  &68.33\% & 52.08\%  &40.83\% &  10.83\% & 43.02\%   \\
\midrule
\rowcolor{gray!20} \textit{Embodied MLLM} & & & & & & \\
RoboBrain-2.0-7B~\citep{robobrain2_0} & &68.33\% &61.25\% &49.58\% &18.33\% &49.38\%\\
RoboBrain-2.5-4B~\citep{robobrain2_5} & &67.36\% &60.67\% &51.05\% &20.50\% &49.90\%\\

 \bottomrule
\end{tabular}
}
\label{table:method ablation}
\vspace{-1em}
\end{table*}

\subsection{Error Analysis \& Qualitative Results}
\label{error_anslysis_qulitative_results}

\begin{figure}[t]
  \centering
  \includegraphics[width=\linewidth]{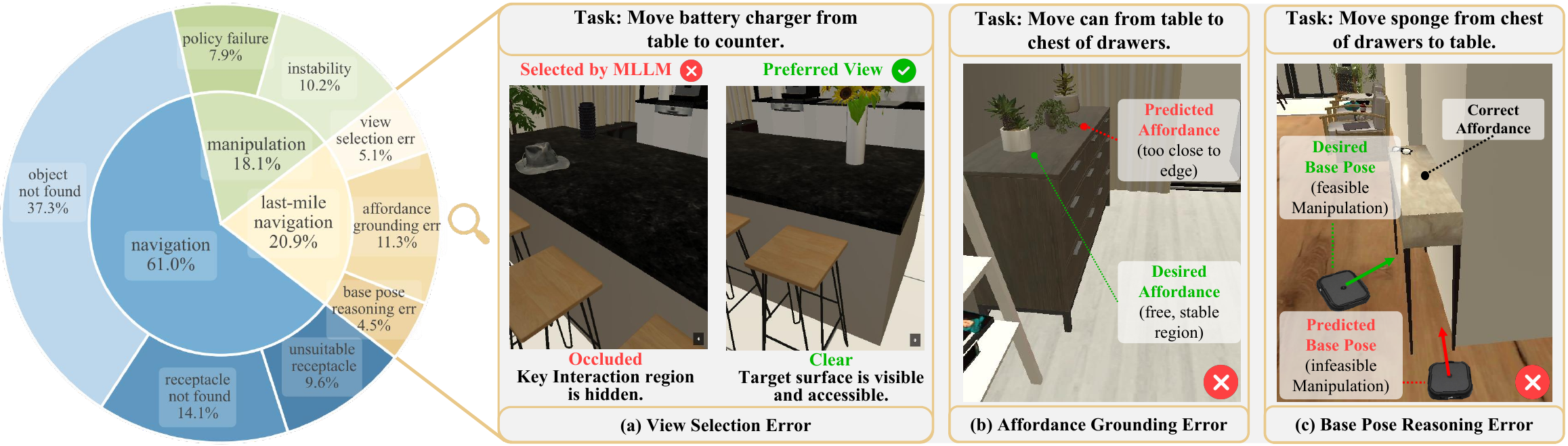}
  \caption{\textbf{Left}: Manual error breakdown on the 20\% OVMM subset used in our ablation studies.
\textbf{Right}: Representative last-mile navigation failures involving view selection, affordance grounding, and base-pose reasoning. The \textcolor{mygreen_error_analysis}{green annotations} indicate preferred views, affordance regions, or base poses, while the \textcolor{myred_error_analysis}{red annotations} show the corresponding erroneous predictions.}
  \label{fig:error_analysis_qual_results}
\end{figure}

To better understand the failure modes of \ourmethod, we manually analyze failed episodes on the subset used in ablation studies, as shown in~\cref{fig:error_analysis_qual_results}. 
We focus on last-mile navigation, which accounts for 20.9\% of failed episodes, and leave navigation and manipulation errors to the Appendix.




Within last-mile navigation, the most frequent failure is affordance grounding error.
As shown in ~\cref{fig:error_analysis_qual_results}(b), the MLLM predicts an affordance point near the boundary of the receptacle, which leads to unstable placement and eventual failure.
The second type of last-mile failure is view selection error. 
Fig.~\ref{fig:error_analysis_qual_results}(a) shows a representative case where the robot's access to the receptacle is blocked by chairs in the selected view, making the receptacle surface difficult to reach and ultimately causing manipulation failure, while another candidate view provides clearer visibility and better manipulation accessibility.
The third type is base-pose reasoning error. As shown in Fig.~\ref{fig:error_analysis_qual_results}(c), the predicted base pose is close to a side wall, making the robot poorly aligned with the target placement point and limiting the manipulator's reachable workspace. In contrast, the desired frontal base pose marked in green provides a more feasible manipulation configuration. Overall, these failures suggest that, despite strong perception and reasoning abilities, current MLLMs still struggle with stage-specific embodied decision making in last-mile navigation.


\subsection{Real-World Experiments}
To validate the real-world performance of  \ourmethod, we deploy it to a Unitree B2 robot~\citep{unitree_b2} with a 6-DoF Unitree Z1 manipulator~\citep{unitree_z1} and Orbbec Gemini 335 camera~\citep{orbbec_gemini335} in an office environment, as shown in \cref{fig:teaser}.
We consider four representative mobile manipulation tasks, whose instructions are shown in ~\cref{tab:real_world_results}.
Each task is repeated ten times from different initial robot poses, simulating diverse terminal states after object navigation toward the table. 

As shown in~\cref{tab:real_world_results}, \ourmethod achieves strong performance on tasks (1) and (2).
Since the two tasks require the robot to identify task-relevant objects and ground an appropriate affordance for interaction, these results indicate that \ourmethod can effectively bridge object-navigation with manipulation in real scenes.
\ourmethod attains lower success rates on task (3) and (4) that require fine-grained task-aware spatial reasoning, such as resolving ``the bottom-left corner of the table''.
These results indicate that MLLMs' spatial reasoning ability can be a bottleneck of \ourmethod.

\begin{table}[t]
\centering
\small
\caption{Real-world task success rate across four representative mobile manipulation tasks.}
\vspace{1mm}
\setlength{\tabcolsep}{4pt}
\begin{tabular}{@{}ccccc@{}}
\toprule
\textbf{\makecell{Pick Cup\\from Table}}
& \textbf{\makecell{Place Cake\\on Plate}}
& \textbf{\makecell{Place Bottle\\in Front of Monitor}}
& \textbf{\makecell{Place Cake\\at Bottom-left Corner}}
& \textbf{Total Success} \\
\midrule
7/10 & 6/10 & 4/10 & 4/10 & 52.5\% \\
\bottomrule
\end{tabular}
\label{tab:real_world_results}
\vspace{-1em}
\end{table}


\section{Conclusion}
\label{sec:Conclusion}

We present \ourmethod, a zero-shot MLLM framework for improving the navigation-to-manipulation handoff in mobile manipulation. By selecting task-relevant views, grounding affordances, and reasoning about manipulation-ready base poses, \ourmethod enables the robot to approach targets in a more physically executable configuration for downstream manipulation. Experiments on the OVMM benchmark show that \ourmethod achieves state-of-the-art Overall SR among compared methods, while ablations validate the contribution of the components. These results highlight the potential of MLLMs as visual-spatial reasoners for last-mile navigation.

\textbf{Limitations}\quad \ourmethod assumes that object navigation can bring the robot to a near-target state where the target object or receptacle appears in recent observations. This assumption can be relaxed in future work by incorporating active local exploration before last-mile navigation. In addition, as shown in~\cref{fig:error_analysis_qual_results}, some failures arise from imperfect view selection, affordance grounding, and base-pose reasoning; we plan to improve these decisions with stronger prompting and rechecking mechanisms. Finally, we will further evaluate \ourmethod on additional OVMM and last-mile navigation benchmarks to better assess its generalization.





\bibliography{reference_header,ref}

\newpage

\setcounter{footnote}{0}

\appendix
\startcontents[appendix] 

\newcommand{\blueprompt}[1]{\textcolor{blue}{\texttt\textbf{{#1}}}}

The Appendix is organized into the following sections:  
\textbf{Method Details} (Section~\ref{Appendix:Method}),
\textbf{More Experiments \& Results} (Section~\ref{Appendix:Experiments and Results}).

\printcontents[appendix]{}{1}{\setcounter{tocdepth}{2}}




\section{Method Details}
\label{Appendix:Method}

\subsection{View Selection}

The prompt designs for pick and place tasks are illustrated in~\cref{fig:view selection prompt}. For both task types, we set the short-term memory buffer size to $K=5$ by default. To make candidate observations easy for the MLLM to distinguish, we overlay an ID number on the bottom-right corner of each image and label the observations from 0 to 4 in temporal order. The MLLM is instructed to return exactly one selected observation ID in JSON format. As specified in the prompt, the selected observation should not only provide clear visibility of the target object or receptacle, but also contain visual evidence of a feasible approach path that allows the robot to navigate close enough for manipulation.
The selected observation is then used as the reference view for subsequent task-conditioned affordance grounding and geometry-aware base-pose reasoning.

\begin{figure}[htbp]
\centering
\begin{minipage}{\linewidth}\vspace{0mm}    \centering
\begin{tcolorbox}
\fontsize{7.0pt}{0.8\baselineskip}\selectfont

\blueprompt{PROMPT FOR PICK}\\
You are given \{nav\_to\_obj\_memory\_size\} first-person RGB images from a robot's camera. Each image has an ID number in the bottom-right corner (IDs: \{image\_id\_list\}). The robot needs to pick up a \{object\_name\}. \\\\
Task: Choose exactly ONE best image ID from \{image\_id\_list\}.\\
Your choice MUST satisfy BOTH criteria below:\\
1) Visibility: The \{object\_name\} is clearly visible and identifiable (not blurry, not heavily occluded, and enough of the object is visible to pick a precise grasp/target point).\\
2) Reachability/Navigability: The robot can realistically move close enough to the \{object\_name\} to pick it up.
Avoid images where obstacles/clutter block the approach path to the \{object\_name\} or where the \{object\_name\} appears unlikely to be reachable.\\

Return a JSON object with key `ID' where `ID' is one of \{image\_id\_list\}. If you cannot find the \{object\_name\}, still return the best-guess image ID for the most likely location of the \{object\_name\}.\\

\blueprompt{PROMPT FOR PLACE}\\
You are given \{nav\_to\_rec\_memory\_size\} first-person RGB images from a robot's camera. Each image has an ID number in the bottom-right corner (IDs: \{image\_id\_list\}). The robot needs to place the \{object\_name\} it is holding onto a \{place\_recep\_name\}. \\

Task: Choose exactly ONE best image ID from \{image\_id\_list\}.\\
Your choice MUST satisfy BOTH criteria below:\\
1) Visibility: The surface of the \{place\_recep\_name\} is clearly visible (not blurry, not heavily occluded, enough surface area visible to select a stable point for placing the object).

2) Reachability/Navigability: The robot can realistically move close enough to the \{place\_recep\_name\} to perform placement. Avoid images where obstacles/clutter block access to the front/edge of the \{place\_recep\_name\} or where approaching the \{place\_recep\_name\} would likely be impossible.\\

Return a JSON object with key `ID' where `ID' is one of \{image\_id\_list\}.
If you cannot find the \{place\_recep\_name\}, still return the best-guess image ID for the most likely location of the \{place\_recep\_name\}

\end{tcolorbox}
\caption{View Selection Prompt for Pick and Place}
\label{fig:view selection prompt}
\end{minipage}
\vspace{-0.5em}
\end{figure}

\subsection{Task-Conditioned Affordance Grounding}

\cref{fig:affordance_grounding_prompt} illustrates the pick and place prompt designs for task-conditioned affordance grounding.
To make the MLLM output resolution-independent and easier to align with the input image, we require the model to predict the affordance grounding result as a normalized 2D coordinate.
The normalized coordinate is then mapped back to the pixel coordinate $(u, v)$ according to the image width and height.
For MLLM backends that are explicitly trained for grounding and adopt their own coordinate conventions, such as Qwen3-VL and RoboBrain-2.5, however, we modify the prompt to follow their native grounding format.
Specifically, these models represent grounding coordinates in a 0--1000 coordinate space, which we subsequently rescale to the original image resolution.

\begin{figure}[htbp]
\centering
\begin{minipage}{\linewidth}\vspace{0mm}    \centering
\begin{tcolorbox}
\fontsize{7.0pt}{0.8\baselineskip}\selectfont
\blueprompt{PROMPT FOR PICK}\\
You are given ONE first-person RGB image from a robot's camera. The robot needs to pick up a \{object\_name\}. \\

Task: Select a pixel (u, v) that points to the \{object\_name\} to be picked up. Requirements (MUST satisfy both):\\
1) Visible-on-object point: The pixel must lie ON the \{object\_name\} (not background or other items), and should be on a clearly visible, unoccluded part of the object (avoid blurry/ambiguous regions and heavy occlusions).\\
2) Reachable for pickup: Choose a point such that the robot can realistically approach near and pick up the object. Prefer points on the object that are not blocked by obstacles between the robot and other object. Avoid cases where the object (or the selected point) is behind large barriers so the robot cannot get close enough to grasp.\\

Return the pixel coordinates (u, v) as a JSON object with keys `u' and `v'. The coordinates must lie strictly within the open interval (0, 1) (i.e., $0 < coordinate < 1$), indicating the normalized pixel locations of the points in the image. Here, u is the horizontal coordinate (0 at the left edge → 1 at the right edge) and v is the vertical coordinate (0 at the top edge → 1 at the bottom edge), i.e., origin at the top-left. If you cannot find the \{object\_name\}, still return your best-guess normalized pixel coordinates (u, v) as a JSON object for the most likely location of the \{object\_name\}.\\

\blueprompt{PROMPT FOR PLACE}\\
You are given ONE first-person RGB image from a robot's camera. The robot is holding an \{object\_name\} and needs to place it on a \{place\_recep\_name\}.\\ 

Task: Select a pixel (u, v) that indicates a good placement point on the surface of the \{place\_recep\_name\}. Requirements (MUST satisfy all):\\
1) Safe \& stable surface: The point must lie on a flat, supported, and stable region of the \{place\_recep\_name\}. The point should NOT overlap/collide with any other unrelated objects.\\
2) Avoid the edge (anti-drop): Prefer an interior placement region rather than the boundary. Do NOT select points on or near the \{place\_recep\_name\}'s edge.\\
3) Reachable placement (base + arm constraints): The point should be reachable for the robot to approach and place the object. Prefer locations with clear free space around the receptacle and no obvious obstacles blocking the robot's approach and placing path (e.g., clutter, furniture blocking the front of the \{place\_recep\_name\}). Avoid points that appear behind barriers or in tightly cluttered regions.\\

Return the pixel coordinates (u, v) as a JSON object with keys `u' and `v'. The coordinates must lie strictly within the open interval (0, 1) (i.e., $0 < coordinate < 1$), indicating the normalized pixel locations of the points in the image. Here, u is the horizontal coordinate (0 at the left edge → 1 at the right edge) and v is the vertical coordinate (0 at the top edge → 1 at the bottom edge), i.e., origin at the top-left. If you cannot find the \{place\_recep\_name\}, still return your best-guess normalized pixel coordinates (u, v) as a JSON object for the most likely location of the \{place\_recep\_name\}.

\end{tcolorbox}
\caption{Affordance Grounding Prompt for Pick and Place}
\label{fig:affordance_grounding_prompt}
\end{minipage}
\vspace{-0.5em}
\end{figure}


\subsection{Geometry-Aware Base-Pose Reasoning}
\cref{fig:base_pose_reasoning_prompt} illustrates our prompt design for geometry-aware base-pose reasoning.
We provide the MLLM with the visually prompted observation, robot configuration, task instruction, and lifted target coordinate.
In addition to predicting target base position, we ask the MLLM to predict the arm extension and lift height required for manipulation, which are used as a simple manipulation policy for \ourmethod.

\begin{figure}[htbp]
\centering
\begin{minipage}{\linewidth}\vspace{0mm}    \centering
\begin{tcolorbox}
\fontsize{7.0pt}{0.8\baselineskip}\selectfont
\blueprompt{PROMPT FOR PICK}\\
You are a mobile manipulator robot planner. Use the ROS coordinate system definition: \\
- X-axis: Forward (Front of the robot)\\
- Y-axis: Left\\
- Z-axis: Up\\
- Theta: Rotation around Z-axis (counter-clockwise is positive, 0 points to X-Forward)\\

Task: Determine an optimal manipulation-ready configuration for the robot to pick up the \{object\_name\}, including both the base pose (x, y, theta) and the arm settings (arm\_reach, arm\_lift).\\

Current Status:\\
The robot is currently at (0, 0, 0) in its local frame.\\
The target is located at: (\{target\_3d\_position[0]\}, \{target\_3d\_position[1]\}, \{target\_3d\_position[2]\}) meters relative to the robot's current center.\\
Interpretation: this 3D target coordinate refers to the \{object\_name\}'s location in the robot coordinate system.\\
As illustrated in the image, the robot's base camera provides a first-person view, and the target location is marked with a red dot.\\

Constraints: \\
Robot Configuration: \{robot\_config\}\\
Optimal Manipulation Distance: The base should typically be placed such that the object is within 70--80\% of the max arm reach.\\
Orientation: The robot should face the object (Target Theta should align X-axis with the object).\\
Collision Avoidance: The chosen base pose and arm settings must avoid collisions with any objects in the scene.\\
For picking task, first estimate the object's height based on its name, then compute arm\_lift as:\\
arm\_lift = target location height + \{object\_name\}'s height / 2 + 0.2 meters for gripper's height, since the gripper typically holds the object around its center.\\            

Output:\\
Return a JSON object with keys `x', `y', `theta', `arm\_reach', and `arm\_lift' representing the `RELATIVE target pose' and arm settings for manipulation.\\
x, y: meters (in current base frame)\\
theta: radians (relative rotation from current heading)\\
arm\_reach: meters, the arm extension length\\
arm\_lift: meters, the arm lift height\\
The values `x', `y', `theta', `arm\_reach', and `arm\_lift' together form the key elements for the robot to complete the pick manipulation.\\

\blueprompt{PROMPT FOR PLACE}\\
You are a mobile manipulator robot planner. Use the ROS coordinate system definition:\\
- X-axis: Forward (Front of the robot)\\
- Y-axis: Left\\
- Z-axis: Up\\
- Theta: Rotation around Z-axis (counter-clockwise is positive, 0 points to X-Forward)\\

Task: Determine an optimal manipulation-ready configuration for the robot to place the held \{object\_name\} at the target location, including both the base pose (x, y, theta) and the arm settings (arm\_reach, arm\_lift).\\

Current Status:\\
The robot is currently at (0, 0, 0) in its local frame.\\
The target is located at: (\{target\_3d\_position[0]\}, \{target\_3d\_position[1]\}, \{target\_3d\_position[2]\}) meters relative to the robot's current center.\\
Interpretation: this target 3D location refers to the receptacle's top surface location in the robot coordinate system.\\
As illustrated in the image, the robot's base camera provides a first-person view, and the target location is marked with a red dot.\\

Constraints:\\
Robot Configuration: \{robot\_config\}\\
Optimal Manipulation Distance: The base should typically be placed such that the object is within 70--80\% of the max arm reach.\\
Orientation: The robot should face the receptacle (Target Theta should align X-axis with the receptacle).\\
Collision Avoidance: The chosen base pose and arm settings must avoid collisions with any objects in the scene.\\
For placement task, first estimate the object's height based on its name, then compute arm\_lift as:\\ 
arm\_lift = target location height + \{object\_name\}'s height / 2 + 0.2 meters for gripper's height, since the gripper typically holds the object around its center.\\         

Output:\\
Return a JSON object with keys `x', `y', `theta', `arm\_reach', and `arm\_lift' representing the `RELATIVE target pose' and arm settings for manipulation.\\
x, y: meters (in current base frame)\\
theta: radians (relative rotation from current heading)\\
arm\_reach: meters, the arm extension length\\
arm\_lift: meters, the arm lift height\\
The values `x', `y', `theta', `arm\_reach', and `arm\_lift' together form the key elements for the robot to complete the place manipulation.

\end{tcolorbox}
\caption{Base-pose Reasoning Prompt for Pick and Place}
\label{fig:base_pose_reasoning_prompt}
\end{minipage}
\end{figure}

\section{More Experiments \& Results}
\label{Appendix:Experiments and Results}

\subsection{Experimental Settings}
We evaluate \ourmethod with Gemini-3-Flash-Preview~\citep{gemini-3-flash}, GPT-series models~\citep{gpt4.1, gpt5.2, gpt5.4}, Qwen3-VL models~\citep{qwen3-vl}, and RoboBrain models~\citep{robobrain2_0, robobrain2_5} as MLLM backends. Unless otherwise specified, all models are used with their default inference settings.

\subsection{Additional Ablations}
All ablations are conducted on the same scene-stratified random 20\% subset of the OVMM benchmark used for the ablation studies in the main paper, due to the high computational cost of embodied evaluation.


\paragraph{Merging View Selection and Affordance Grounding.}
We ablate the explicit decomposition of view selection and affordance grounding to examine whether the MLLM can jointly perform these two stages in a single call.
Instead of first selecting one image from the five observations in the short-term memory buffer and then grounding an affordance point in the selected image, we provide all five candidate images to the MLLM simultaneously and ask it to output both the selected image ID and the corresponding affordance-point coordinate. The downstream 3D lifting, base-pose reasoning remain unchanged.

As shown in~\cref{table:merge_view_selection_affordance_grounding}, merging view selection and affordance grounding leads to overall performance drops for Qwen3-VL-8B-Instruct, Qwen3-VL-32B-Instruct, and Gemini-3-Flash-Preview.
Although these two steps are closely related and may appear natural for humans to perform jointly, the results suggest that current MLLMs still struggle to reliably handle view selection and precise affordance grounding within a single reasoning step.
This further supports our design choice of explicitly decomposing last-mile navigation into simpler MLLM calls, allowing each call to better leverage the model's perception and reasoning capabilities for the corresponding subtask.

\begin{table*}[t]

\caption{Ablation on merging view selection and affordance grounding, on the 20\% OVMM subset. $\Delta$ denotes the change in success rate relative to the corresponding base configuration with the same MLLM backend, i.e., \colorbox{blue!10}{Gemini-3-Flash-Preview}, \colorbox{yellow!10}{Qwen3-VL-8B-Instruct}, and \colorbox{green!10}{Qwen3-VL-32B-Instruct}. The reported results validate the effectiveness of \ourmethod's explicit decomposition of view selection and affordance grounding, which yields higher Overall SR across the evaluated MLLM backends.}

\centering
\small
\resizebox{\linewidth}{!}{
\begin{tabular}{@{}lcccccc@{}}
\toprule
\multirow{2}{*}{\textbf{Method}}
& & \multicolumn{3}{c}{\textbf{Partial Success Rates}} &\multirow{2}{*}{\textbf{\makecell{Overall
\\ SR}}} &\multirow{2}{*}{\textbf{\makecell{Average
\\ SR}}} \\
\cmidrule{3-5} 
  & & FindObj~($\uparrow$) & Pick~($\uparrow$) & FindRec~($\uparrow$)  \\
 \midrule
\rowcolor{blue!10}
\ourmethod w/ Gemini-3-Flash-Preview~\citep{gemini-3-flash} & &67.91\% &65.83\% &54.58\% &25.42\% &53.43\% \\
\rowcolor{blue!10}
$\Delta$ merging view selection \& affordance grounding & & \textcolor{blue}{$0.42\%\uparrow$} &  \textcolor{red}{$7.08\%\downarrow$} &  \textcolor{red}{$5.00\%\downarrow$} &  \textcolor{red}{$4.17\%\downarrow$} &  \textcolor{red}{$3.95\%\downarrow$} \\
\midrule
\rowcolor{yellow!10}
\ourmethod w/ Qwen3-VL-8B-Instruct~\citep{qwen3-vl} & &68.75\%  &60.00\%  &47.92\%  &9.58\%  &46.56\%  \\
\rowcolor{yellow!10}
$\Delta$ merging view selection \& affordance grounding & & \textcolor{red}{$0.42\%\downarrow$} & \textcolor{red}{$5.42\%\downarrow$} & \textcolor{red}{$4.17\%\downarrow$} & \textcolor{red}{$1.66\%\downarrow$} & \textcolor{red}{$2.91\%\downarrow$} \\
\midrule
\rowcolor{green!10}
\ourmethod w/ Qwen3-VL-32B-Instruct~\citep{qwen3-vl} &  &67.92\% &62.08\%  &52.08\% &  15.83\% & 49.48\%   \\
\rowcolor{green!10}
$\Delta$ merging view selection \& affordance grounding & & \textcolor{red}{$0.42\%\downarrow$} &  \textcolor{red}{$4.58\%\downarrow$} &  \textcolor{red}{$4.16\%\downarrow$} &  \textcolor{red}{$4.16\%\downarrow$} &  \textcolor{red}{$3.33\%\downarrow$} \\
 \bottomrule
\end{tabular}
}
\label{table:merge_view_selection_affordance_grounding}
\vspace{-1em}
\end{table*}

\paragraph{Geometrically Computed Heading vs. MLLM-Predicted Heading.}
In our full method, instead of relying on the MLLM to predict the heading angle, we geometrically compute the final orientation by turning the robot toward the lifted affordance point.
To evaluate this design choice, we conduct a heading-strategy ablation comparing the MLLM-predicted heading with our geometrically computed heading, as shown in~\cref{table:geometric_computed_heading_MLLM_predicted_heading}. 

The results show that geometrically computing the heading consistently outperforms directly using the MLLM-predicted heading.
While the two variants achieve similar FindRec success rates, the geometrically computed heading leads to a substantially higher Overall SR than the MLLM-predicted heading.
This suggests that directly predicting the heading angle remains challenging for current MLLMs, whereas orienting the robot toward the lifted affordance point better constrains the final base pose and improves downstream manipulation success.

\begin{table*}[t]

\caption{Heading-strategy ablation on the OVMM benchmark (20\% subset) using Gemini-3-Flash-Preview as the MLLM backend.
We compare our geometrically computed heading, which orients the robot toward the lifted affordance point, with directly using the heading angle predicted by the MLLM during base-pose reasoning.}
\centering
\small
\resizebox{\linewidth}{!}{
\begin{tabular}{@{}lcccccc@{}}
\toprule
\multirow{2}{*}{\textbf{Method}}
& & \multicolumn{3}{c}{\textbf{Partial Success Rates}} &\multirow{2}{*}{\textbf{\makecell{Overall
\\ SR}}} &\multirow{2}{*}{\textbf{\makecell{Average
\\ SR}}} \\
\cmidrule{3-5} 
  & & FindObj~($\uparrow$) & Pick~($\uparrow$) & FindRec~($\uparrow$)  \\
 \midrule
 \ourmethod w/ geometrically computed heading & &67.91\% &65.83\% &54.58\% &25.42\% &53.43\% \\
\ourmethod w/ MLLM predicted heading & &67.50\% &63.33\% &52.92\% &17.08\% &50.21\%\\
 \bottomrule
\end{tabular}
}
\label{table:geometric_computed_heading_MLLM_predicted_heading}
\end{table*}

\begin{table*}[h]
\caption{Ablation study on thinking vs. non-thinking models on the 20\% OVMM subset.
Inst. denotes Qwen3-VL-8B-Instruct, Think. denotes Qwen3-VL-8B-Thinking, and CoT denotes adding an explicit thinking prompt.}
\vspace{1mm}
\centering
\small
\resizebox{\linewidth}{!}{
\begin{tabular}{@{}cccccccc@{}}
\toprule
\multicolumn{3}{c}{\textbf{Framework Component}} &
\multicolumn{3}{c}{\textbf{Partial Success Rates}} & \multirow{2}{*}{\textbf{\makecell{Overall
\\ SR}}} &\multirow{2}{*}{\textbf{\makecell{Average
\\ SR}}} \\
\cmidrule(r{0.4em}){1-3} \cmidrule(l{0.4em}){4-6}
view selection & affordance grounding & base-pose reasoning &
FindObj~($\uparrow$) & Pick~($\uparrow$) & FindRec~($\uparrow$) & & \\
\midrule

Inst. & Inst. & Inst. & 68.75\%  &60.00\%  &47.92\%  &9.58\%  &46.56\%  \\
Think. & Inst. & Inst. &  67.92\% & 61.67\% & 50.00\%  & 10.00\% & 47.40\% \\
Inst. & Think. & Inst. & 67.92\%  &  55.83\% & 45.83\% & 5.83\% &  43.85\% \\ 
Inst. & Inst. & Think. & 69.17\%  & 62.92\% & 52.50\% & 20.83\% & 51.35\% \\
Think. & Think. & Think. & 68.75\%  &62.08\%  &51.25\%  &21.25\%  &50.83\%  \\
Inst. & Inst.  & Inst. + CoT  & 69.17\%  & 60.00\%  & 50.00\% & 9.17\% & 47.08\% \\

 \bottomrule
\end{tabular}
}
\label{table:appendix_thinking_vs_non-thinking}
\end{table*}




\paragraph{Thinking vs. Non-Thinking Models.}
\cref{table:appendix_thinking_vs_non-thinking} reports the results of using Qwen3-VL-8B-Instruct and Qwen3-VL-8B-Thinking as the MLLM backends for different components of our framework, including view selection, affordance grounding, and base-pose reasoning.

The results show that, among single-component replacements, replacing the instruct model with the thinking model in the \textbf{base-pose reasoning} module brings the most significant improvement in Overall SR, from 9.58\% to 20.83\%.
This suggests that this stage requires stronger spatial reasoning and embodied task understanding.
In contrast, simply adding an explicit CoT prompt to the instruct model for base-pose reasoning does not improve Overall SR, decreasing it from 9.58\% to 9.17\%.
This indicates that merely prompting the model to think is insufficient for this stage.
Using the thinking model for view selection leads to a slight improvement, while replacing the affordance grounding module with the thinking model degrades performance.
This suggests that additional reasoning is not always beneficial for relatively direct perception-oriented stages such as affordance grounding.

\subsection{More Qualitative Results}

\begin{figure}
    \centering
    \includegraphics[width=\linewidth]{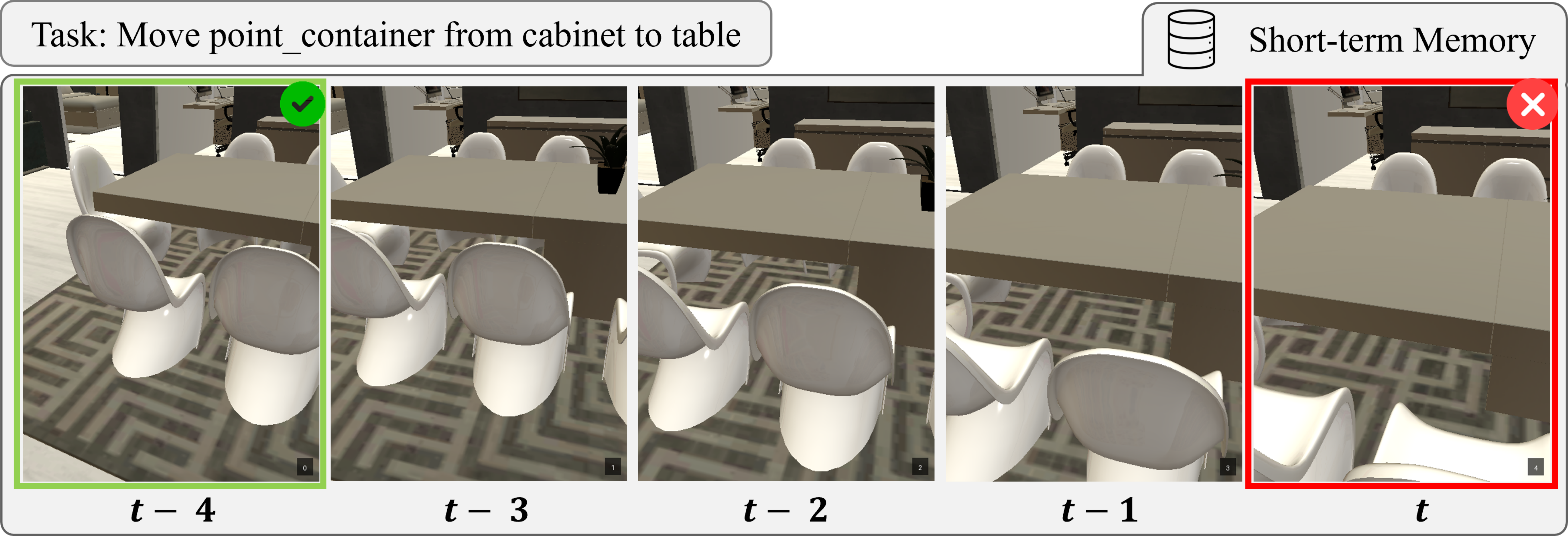}
    \caption{Qualitative example of view selection from the short-term memory $\mathcal{M}_t$. During navigation to the target receptacle, i.e., the table, $\mathcal{M}_t$ stores the most recent five observations. The red box marks the final observation at time $t$, where the robot's approach to the table is blocked by chairs. The green box marks the observation selected by the MLLM, i.e., timestep $t-4$ with ID 0, which provides a better reference view for placement.}
    \label{fig:good_case_view_selection}
\end{figure}

To demonstrate the effectiveness of \ourmethod, we provide successful cases on the OVMM benchmark.
As shown in~\cref{fig:good_case_view_selection}, the observation at the end of navigation to the table is not necessarily the most suitable reference for placement.
In this view, directly executing placement would be obstructed by the chairs in front of the robot, and the limited observation of the surrounding table layout provides insufficient spatial context for further pose adjustment.
By considering the short-term memory $\mathcal{M}_t$, the MLLM selects the observation at timestep $t-4$ (ID 0), which reveals free space near the lower-left side of the table and supports an unobstructed approach to the tabletop, making it a more suitable reference for placement.

\begin{figure*}[t]
  \centering

  \begin{minipage}[t]{0.48\linewidth}
    \vspace{0pt}
    \centering
    \includegraphics[width=\linewidth]{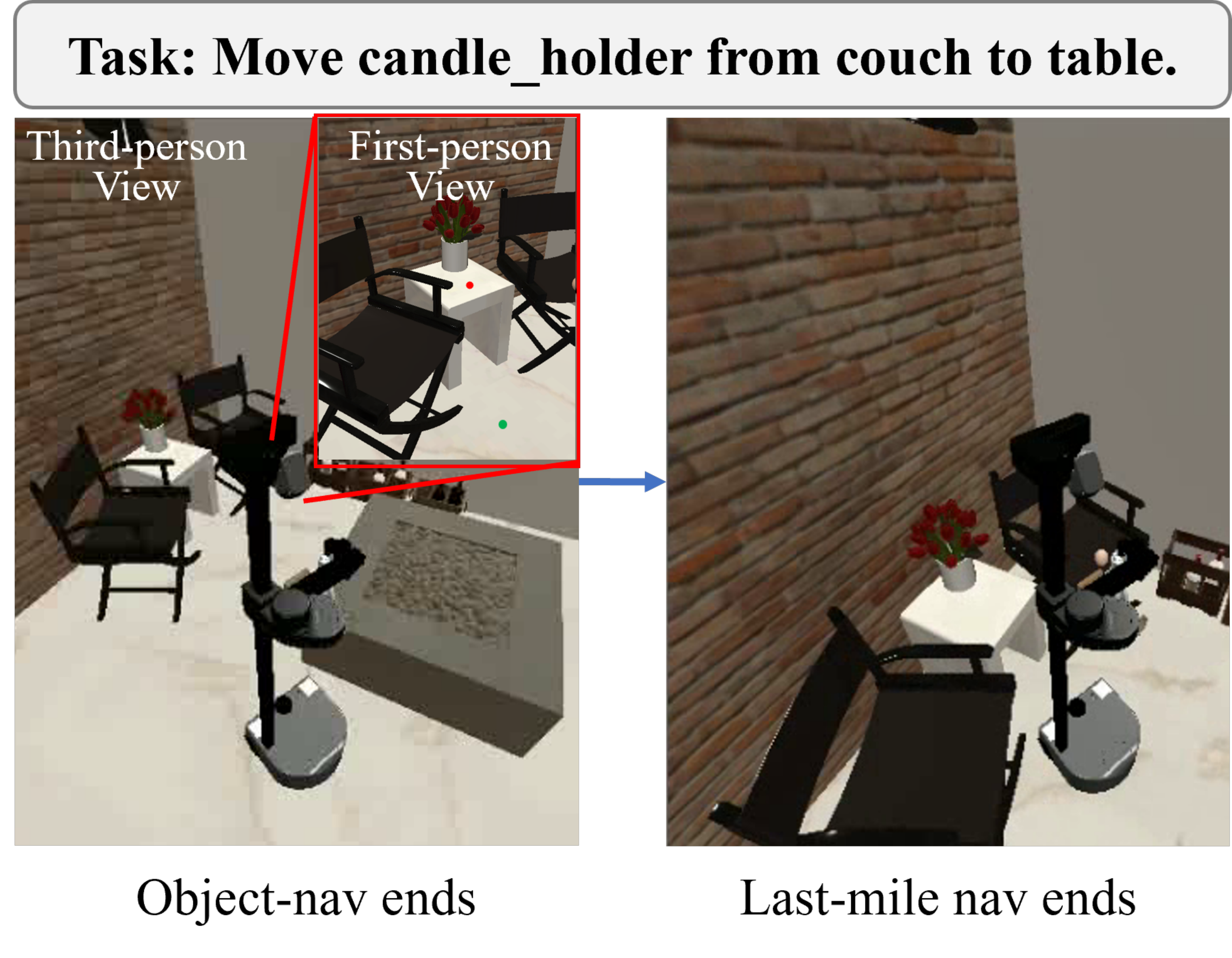}
    \captionof{figure}{Qualitative success case on the OVMM benchmark. After object navigation ends, \ourmethod predicts the affordance point in the first-person view, shown as the \textcolor{myred_error_analysis}{red point}, and infers the robot's base position conditioned on the grounded affordance, task instruction, and robot geometry, shown as the \textcolor{mygreen_error_analysis}{green point}. By executing last-mile navigation based on the predicted base position, the robot successfully adjusts to a manipulation-ready pose.}
    \label{fig:good_case_affordance_grounding_base_pose_reasoning}
  \end{minipage}
  \hfill
  \begin{minipage}[t]{0.48\linewidth}
    \vspace{0pt}
    \centering
    \includegraphics[width=\linewidth]{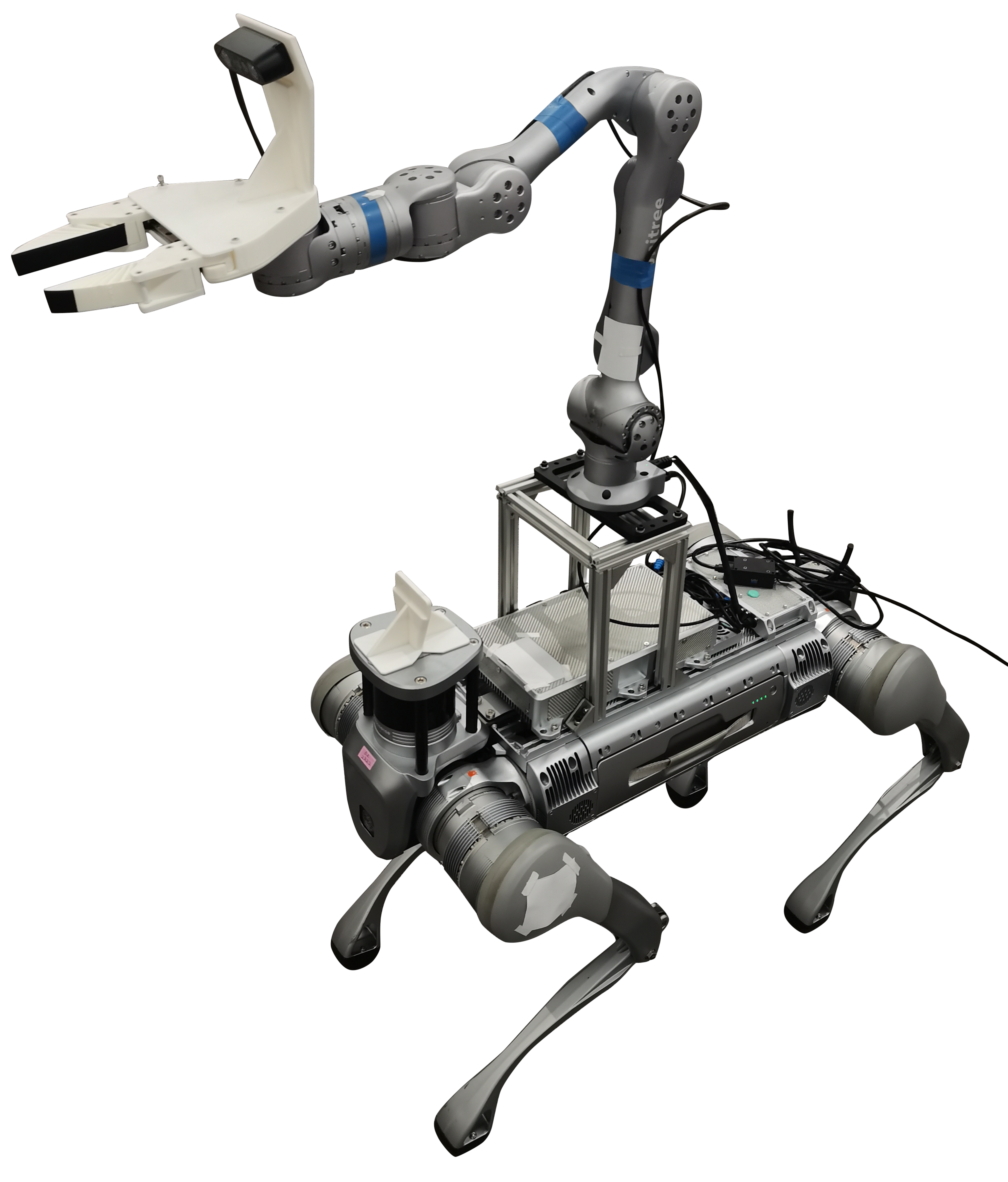}
    \captionof{figure}{Real-world robot with a Unitree B2 robot dog~\citep{unitree_b2} as the mobile base and a 6-DoF Unitree Z1 robotic arm~\citep{unitree_z1} mounted on top used in our experiments.}
    \label{fig:robot_illustration}
  \end{minipage}

\end{figure*}

Furthermore, \cref{fig:good_case_affordance_grounding_base_pose_reasoning} shows another successful case in which the MLLM effectively performs both affordance grounding and base-pose reasoning. After object navigation ends, \ourmethod executes last-mile navigation based on the predicted base pose and successfully adjusts the robot to a manipulation-ready pose.



\subsection{Error Analysis: Navigation}
\label{Appendix_error_analysis_navigation}

As illustrated in~\cref{fig:error_analysis_qual_results} of our main paper, navigation errors account for the largest portion of failures, reaching 61.0\%.
These errors occur before last-mile navigation and mainly include cases where the robot fails to find the target object, fails to find the target receptacle, or navigates to an unsuitable receptacle. In this context, an unsuitable receptacle refers to a semantically correct receptacle whose placement surface is physically infeasible to access due to the robot's embodiment constraints, such as base height, arm reach, or manipulation workspace. For example, although a cabinet surface may be a valid receptacle for the target object, it may be too high for the robot to place the object on it. The high proportion of failures occurring before last-mile navigation suggests that object-goal navigation remains a major bottleneck in OVMM.

\subsection{Error Analysis: Manipulation}
\label{Appendix_error_analysis_Manipulation}

Manipulation errors account for 18.1\% of the failed episodes.
These failures occur after the robot reaches the predicted base pose and mainly fall into two categories: policy failures and placement instability.
Policy failures refer to cases where the low-level manipulation policy produces an inappropriate manipulation action. A typical example is releasing the object from an excessive height, which causes the object to fall or bounce instead of being placed stably on the target surface.

Placement instability refers to cases where the object appears visually static on the receptacle after release but fails to satisfy the strict stability criterion of the OVMM benchmark. Specifically, the benchmark requires the placed object to remain stable under predefined thresholds on linear velocity, angular velocity, and duration. Therefore, an episode can be judged as failed if the object retains small residual motion that is not obvious from visual inspection.

\subsection{Implementation of Real-World Experiments}

For the real-world experiments, we build a quadruped mobile manipulator with a Unitree B2 robot dog~\citep{unitree_b2} as the mobile base and a 6-DoF Unitree Z1 robotic arm~\citep{unitree_z1} mounted on top, as shown in~\cref{fig:robot_illustration}. To accommodate the wrist camera, we replace the official Unitree Z1 two-finger gripper with a custom 3D-printed parallel-jaw gripper that integrates a top-mounted bracket for an Orbbec Gemini 335 RGB-D camera~\citep{orbbec_gemini335}. In this eye-in-hand setup, the camera provides the primary visual input for both navigation and manipulation. We use the quadruped's onboard LiDAR sensors and a LiDAR-inertial SLAM pipeline based on LIO-SAM~\citep{liosam} for occupancy mapping and odometry estimation in the deployment environment. We adopt Gemini-3-Flash-Preview as the MLLM backend in our framework.

For last-mile navigation, we use the ROS~2 Navigation2 stack~\citep{nav2} to drive the robot to the predicted manipulation-ready base poses. 
For real-world picking and placing, we perform an additional affordance refinement step before manipulation. Although the task-conditioned affordance grounding stage provides an initial target point, the projected point may become misaligned in the final manipulation view due to residual odometry errors, depth noise, and calibration errors. After the robot reaches the predicted base pose, we query the MLLM backend again with the current wrist-camera image to revise the manipulation target. For picking, the MLLM grounds a graspable point on the target object and infers the corresponding grasp pose. For placing, it selects a safe and unoccupied point on the target receptacle.



\begin{figure}
    \centering
    \includegraphics[width=\linewidth]{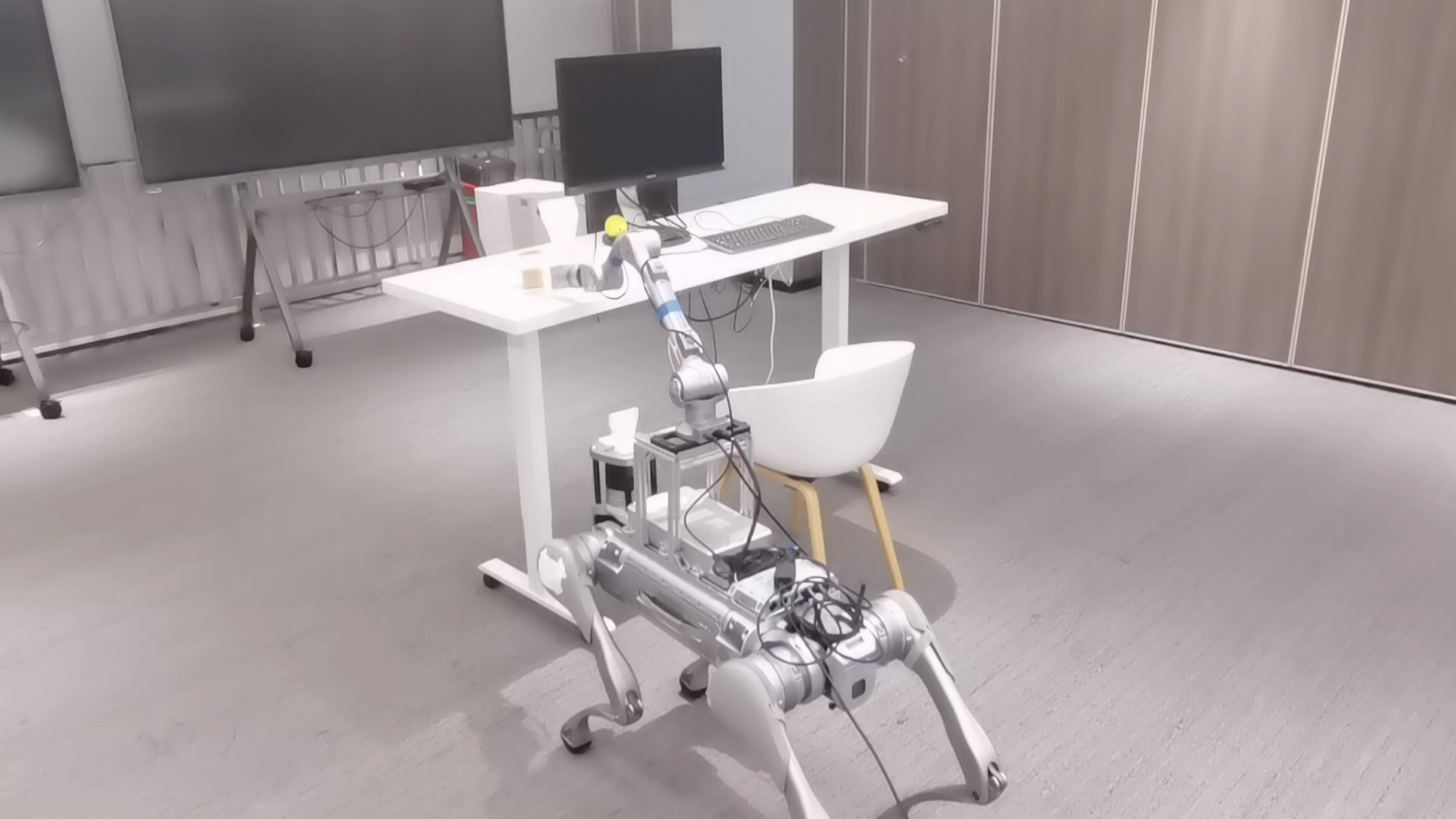}
    \caption{Real-world experiment: Pick up the cup from the table.}
    \label{fig:pick_up_cup}
\end{figure}

\begin{figure}
    \centering
    \includegraphics[width=\linewidth]{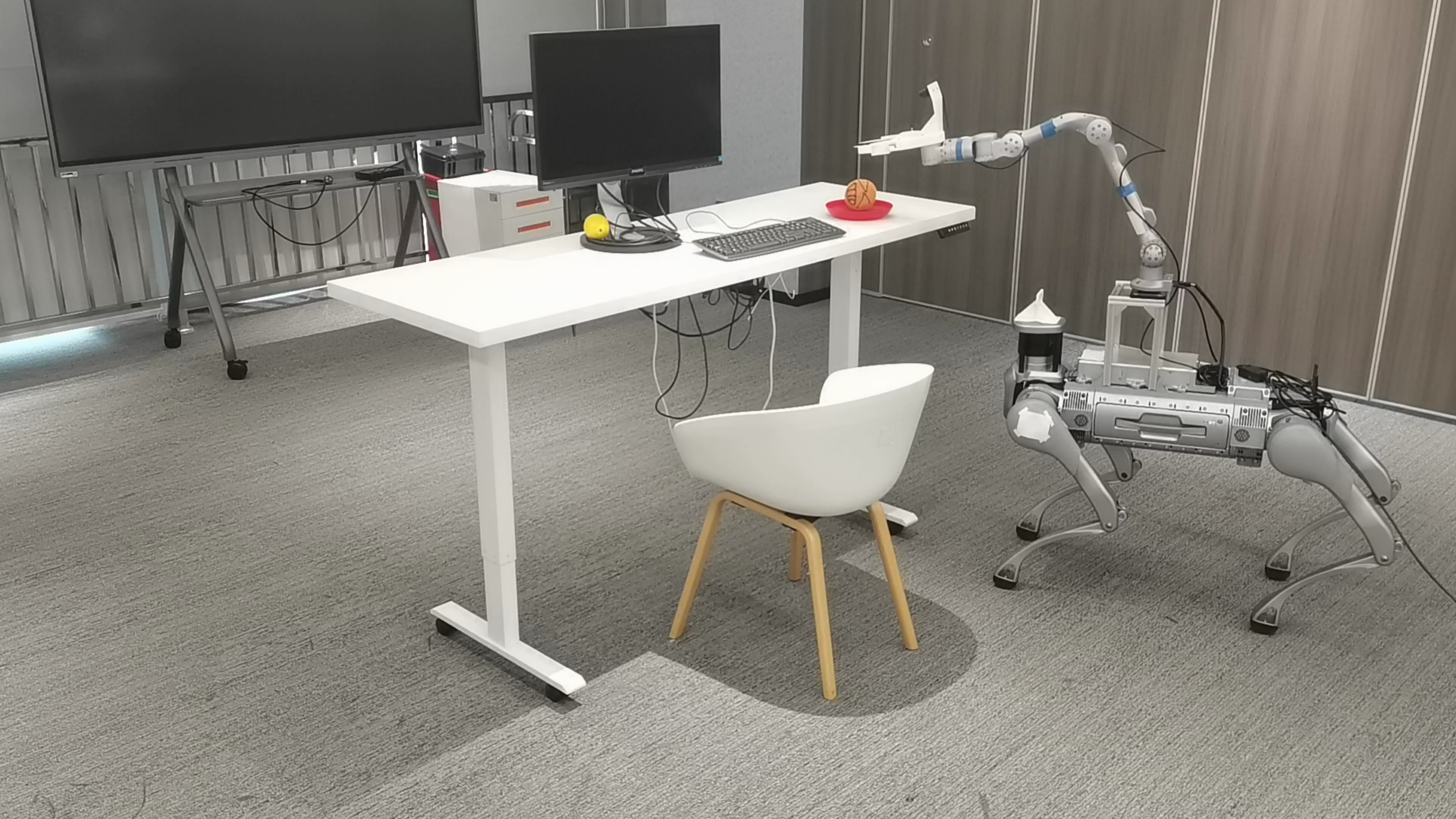}
    \caption{Real-world experiment: Place the cake on the plate.}
    \label{fig:place_cake_on_plate}
\end{figure}

\begin{figure}
    \centering
    \includegraphics[width=\linewidth]{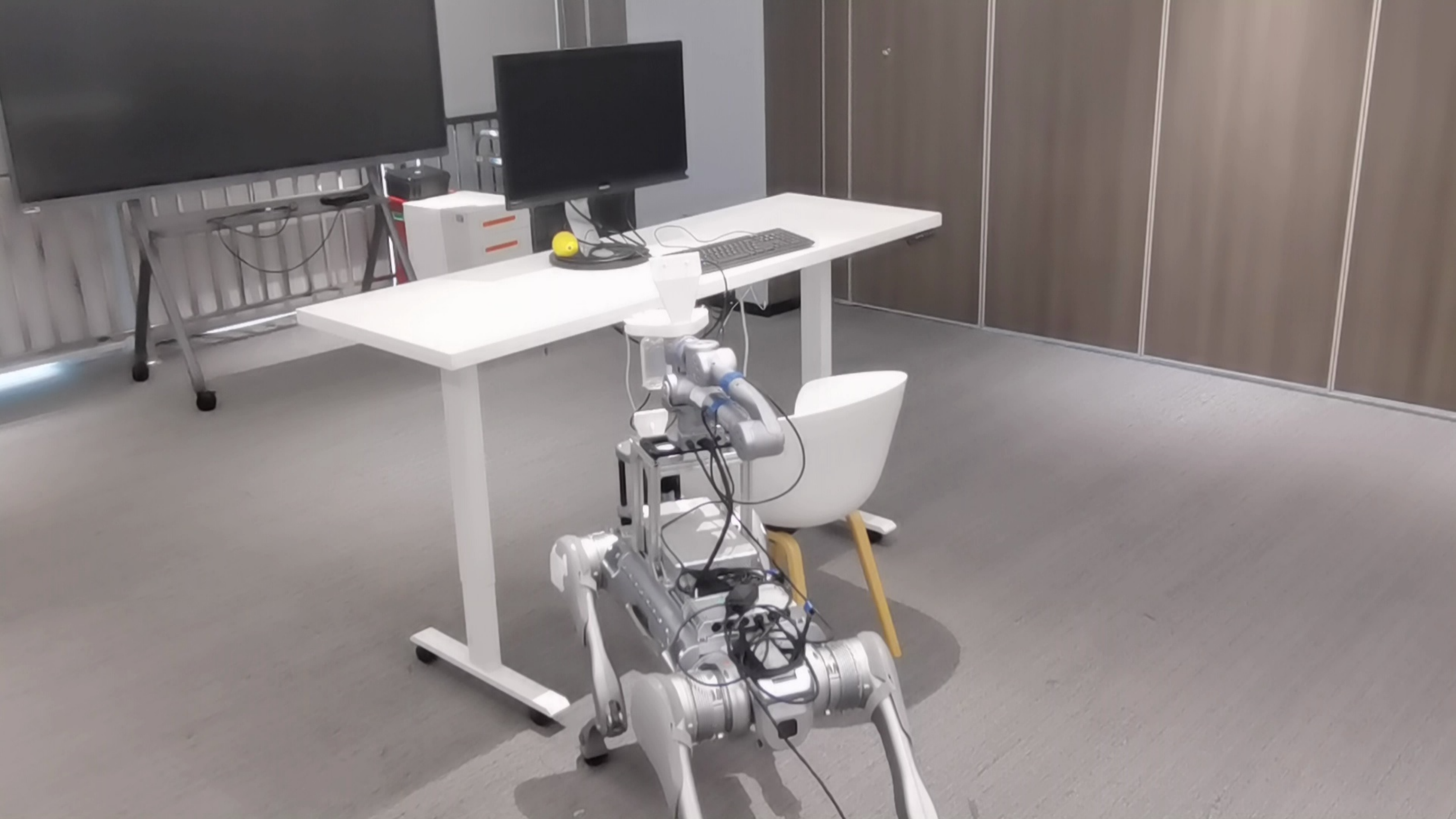}
    \caption{Real-world experiment: Place the bottle on the table, in front of the monitor.}
    \label{fig:place_bottle_in_front_of_monitor}
\end{figure}

\begin{figure}
    \centering
    \includegraphics[width=\linewidth]{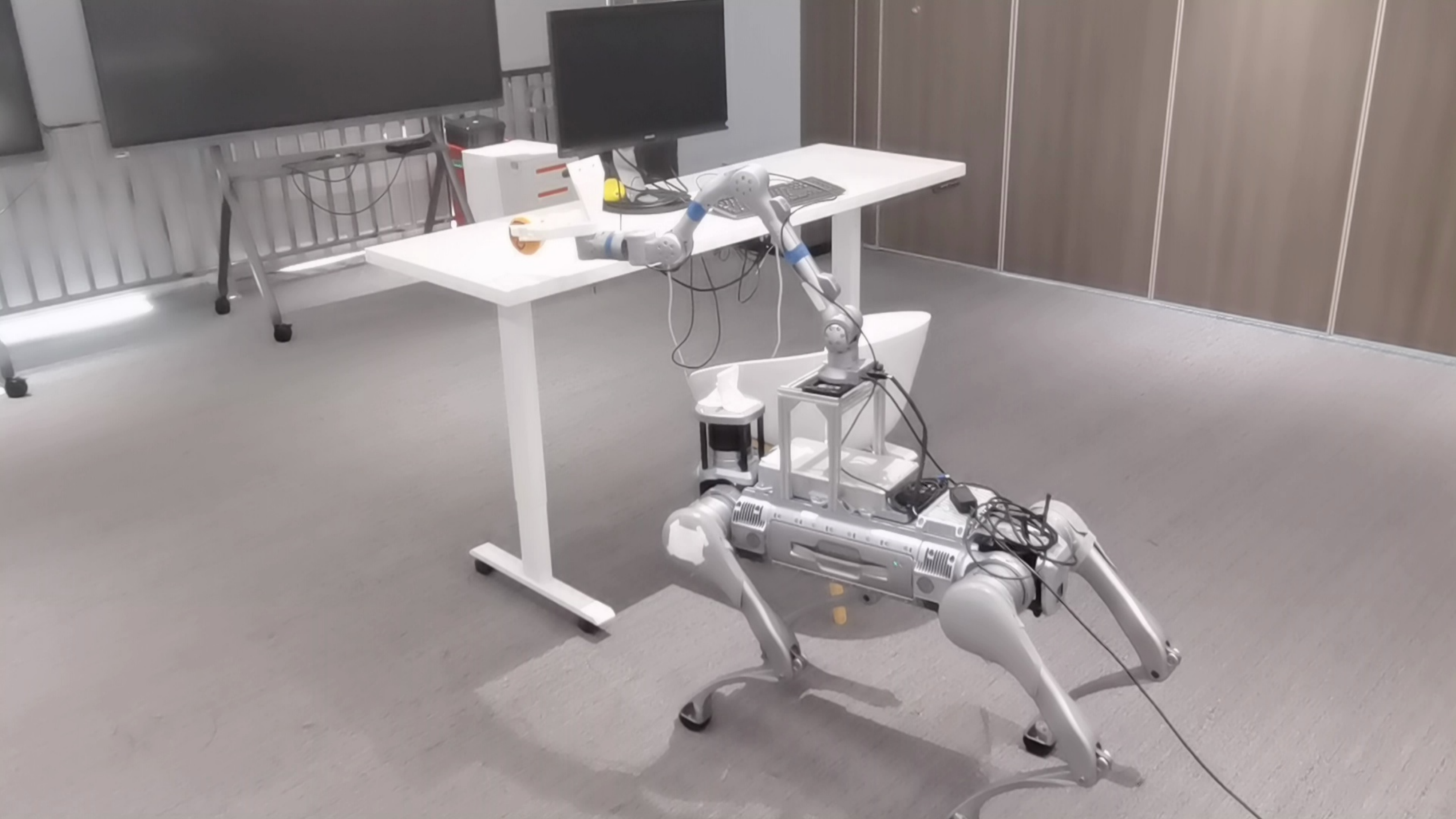}
    \caption{Real-world experiment: Imagine facing the monitor, place the cake at the bottom-left corner of the table.}
    \label{fig:place_cake_bottom_right_corner}
\end{figure}

We conduct four real-world experiments in an office environment: 
\begin{enumerate}
    \item Pick up the cup from the table. (\cref{fig:pick_up_cup})
    \item Place the cake on the plate. (\cref{fig:place_cake_on_plate})
    \item Place the bottle on the table, in front of the monitor. (\cref{fig:place_bottle_in_front_of_monitor})
    \item Imagine facing the monitor, place the cake at the bottom-left corner of the table. (\cref{fig:place_cake_bottom_right_corner})
\end{enumerate}
The first two tasks mainly evaluate whether \ourmethod can effectively bridge object navigation and manipulation in real scenes, where the robot needs to approach the task-relevant object or receptacle and execute the corresponding manipulation.
Furthermore, the latter two tasks focus on fine-grained task-aware spatial reasoning, requiring \ourmethod to resolve spatial relations such as ``in front of the monitor'' and viewpoint-conditioned regions such as ``the bottom-left corner of the table'' before predicting a manipulation-ready base pose. 

\end{document}